\documentclass[runningheads,a4paper]{llncs}

\usepackage{amssymb}
\usepackage{graphicx}

\usepackage{url}

\begin{document}

\mainmatter  

\title{Robust Face Recognition with Deeply Normalized Depth Images}

\titlerunning{Robust Face Recognition with Deeply Normalized Depth Images}

%
%
\author{Ziqing Feng, Qijun Zhao} 
\authorrunning{Ziqing Feng, Qijun Zhao}

\institute{National Key Laboratory of Fundamental Science on Synthetic Vision,\\
College of Computer Science, Sichuan University, Chengdu, China\\
\email{qjzhao@scu.edu.cn}
}

%
%

\maketitle

\begin{abstract}
Depth information has been proven useful for face recognition. However, existing depth-image-based face recognition methods still suffer from noisy depth values and varying poses and expressions. In this paper, we propose a novel method for normalizing facial depth images to frontal pose and neutral expression and extracting robust features from the normalized depth images. The method is implemented via two deep convolutional neural networks (DCNN), normalization network ($Net_{N}$) and feature extraction network ($Net_{F}$). Given a facial depth image, $Net_{N}$ first converts it to an HHA image, from which the 3D face is reconstructed via a DCNN. $Net_{N}$ then generates a pose-and-expression normalized (PEN) depth image from the reconstructed 3D face. The PEN depth image is finally passed to $Net_{F}$, which extracts a robust feature representation via another DCNN for face recognition. Our preliminary evaluation results demonstrate the superiority of the proposed method in recognizing faces of arbitrary poses and expressions with depth images. 

\end{abstract}

\section{Introduction}

Face recognition can be implemented using different modalities of face data, such as RGB images (2D) \cite{Liu_CVPR_2017}, depth images (2.5D) \cite{Lee_BMVC_2016} and shapes (3D) \cite{Liu_CVPR_2018}. Since the advent of low-cost depth sensors (e.g., Kinect and RealSense), depth information has been increasingly used in many applications, including face recognition, semantic segmentation, and robot navigation, etc. It appears that depth images contain rich information that is worth human beings to explore.

RGB-image-based face recognition technology has developed rapidly in the past decade thanks to the emerging deep learning techniques \cite{Alex_NIPS_2012}. However, depth-image-based face recognition still confronts many challenging problems, e.g., noisy depth values, occlusion, pose and expression variations. Some researchers \cite{Lee_BMVC_2016} propose to fuse multiple frames of depth images to improve the precision. Other researchers directly transform depth images to 3D point clouds with an assumed camera model \cite{Zhang_IEEE_2016}, and rotate the faces to frontal pose and complete the invisible areas on the faces via symmetric filling \cite{BiLi_2013}. Although these methods achieve promising results on several benchmark databases, none of them can well cope with facial depth images with both noise in depth values and variations in pose and expression, which frequently occur in real-world applications.

This paper aims to improve the depth-image-based face recognition accuracy particularly on noisy data with varying poses and expressions. To this end, we propose a deep learning (DL) based facial depth image normalization approach. Given a facial depth image of arbitrary view and expression, we first reconstruct its 3D face (represented by a 3D morphable model (3DMM) \cite{Vetter_TPAMI_2003}) via regressing its 3DMM-based shape and expression parameters, then generate a pose-and-expression-normalized (PEN) depth image for the face, and finally use the PEN depth image for face recognition. Preliminary evaluation results show that our proposed method is more robust to depth noise and pose and expression variations than counterpart methods, and thus obtains better face recognition accuracy.

The rest of this paper is organized as follows. Section 2 reviews related work on depth-image-based face recognition. Section 3 introduces in detail our proposed method, and Section 4 reports the experimental results. Section 5 finally concludes the paper.

\section{Related Work}

Existing depth-image-based face recognition methods can be roughly divided into two categories depending on whether 3D faces (as point clouds) are used. Methods in the first category either assume that 3D faces are available during acquisition \cite{BiLi_2013,Berretti_IEEE_2014} or reconstruct 3D faces from depth images or videos \cite{Zhang_IEEE_2016,Sang_CIN_2015}. With the 3D faces, Li et al. \cite{BiLi_2013} and Sang et al. \cite{Sang_CIN_2015} correct the facial pose to frontal, and further preprocess the obtained frontal facial depth images by symmetric filling and smooth resampling. They do not explicitly process varying facial expressions, but require multiple depth images per subject with different expressions in the gallery. Zhang et al. \cite{Zhang_IEEE_2016} directly match the 3D faces by using the iterative closest point algorithm. However, they have to capture videos of 3D faces so that higher-precision 3D faces can be reconstructed to assure good face recognition accuracy. Unlike these methods, our proposed method reconstructs 3D faces from only single depth images, and can remove both pose and expression variations on the faces, resulting in depth images with frontal pose and neutral expression, which are preferred by face recognition systems.

Methods in the second category directly work on depth images, and focus on devising effective feature descriptors and classifiers. A variety of hand-crafted descriptors have been introduced for depth-image-based face recognition, such as local binary patterns \cite{Aissaoui_ICIP_2014}, local quantized patterns \cite{Mantecon_2015}, and bag of dense derivative depth patterns \cite{Mantecón_SPL_2016}. After feature descriptors are extracted from depth images, different classifiers like support vector machines \cite{Olegs_svm_2014} and nearest neighbor classifiers \cite{Goswami_IBTAS_2014} are applied to recognize the faces in the depth images. These methods often suffer from noisy depth values or varying poses and expressions. In this paper, instead of employing hand-crafted feature descriptors, we adopt a deep convolutional neural network (DCNN) to extract features from the normalized facial depth images.

%

\begin{figure}
\setlength{\belowcaptionskip}{-0.1cm}
\centering
\includegraphics[scale=0.23]{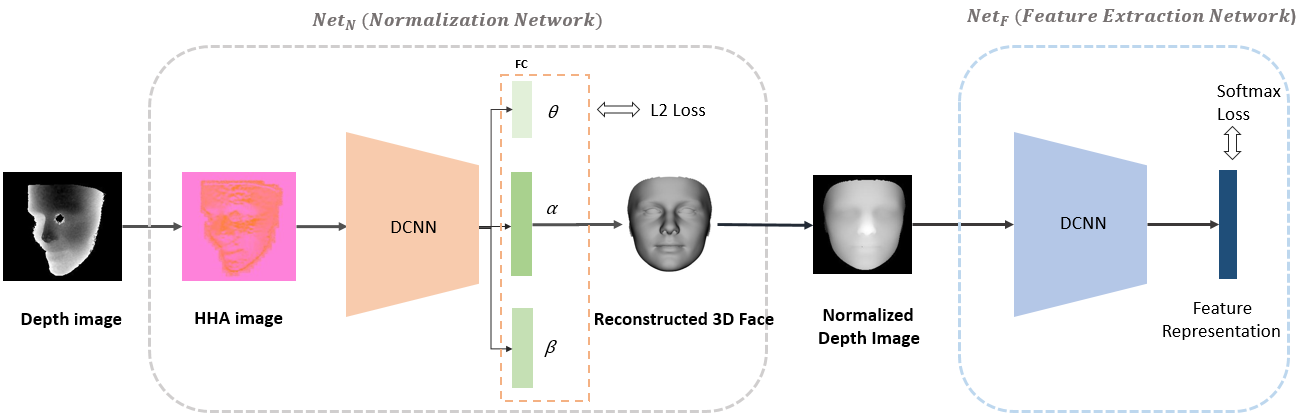}
\caption{Flowchart of our proposed face recognition method using deeply normalized depth images. DCNN denotes deep convolutional neural network. FC denotes fully-connected layer. $\theta$, $\alpha$ and $\beta$ are, respectively, pose, shape and expression parameters of the face in the input depth image.}
\label{fig:flowchart}
\end{figure}

\section{Proposed Method}

\subsection{Overview}
As shown in Fig. \ref{fig:flowchart}, our proposed method consists of two main components, normalization network ($Net_{N}$) and feature extraction network ($Net_{F}$). Given a facial depth image with arbitrary pose and expression, $Net_{N}$ first converts it to an HHA image, from which the 3D face is reconstructed as pose, shape and expression parameters defined by the 3D morphable model (3DMM). $Net_{N}$ then generates from the obtained 3DMM shape parameters a pose-and-expression normalized (PEN) facial depth image that has frontal pose and neutral expression. During this normalization process, invisible facial regions are completed, and noisy depth values can also be regularized, resulting in higher-precision complete facial depth images. Robust features are finally extracted from the normalized depth images by $Net_{F}$, which is trained for face recognition purpose.

\subsection{Normalization Network}
\subsubsection{Data Transformation}
Motivated by RGB-D semantic segmentation methods \cite{Gupta_ECCV_2014}, we transform depth images to HHA images. While depth images, as single-channel images, record only the depth values of pixels, HHA images represent pixels in three channels, including horizontal disparity, height over ground, and the angle between local surface normal and gravity direction. Therefore, HHA images convey more geometric information, which is beneficial to the reconstruction of 3D face shapes. In this paper, we employ the method in \cite{Gupta_ECCV_2014} for this data transformation from depth images to HHA images.

\subsubsection{3D Face Reconstruction}
Let $S=[x_{1}, y_{1}, z_{1}, \cdots, x_{n}, y_{n}, z_{n}]^{T}$ be a 3D face shape of $n$ vertices, where $(x_{i}, y_{i}, z_{i})$ are coordinates of the $i^{\texttt{th}}$ vertex and `$T$' denotes the transpose operator. Following the 3DMM definition, we represent 3D face shapes as
\begin{equation}\label{eq:3dmm}
S = \bar{S} + \Sigma_{k=1}^{K}\alpha_{k}S_{k} + \Sigma_{l=1}^{L}\beta_{l}S_{l}^{Exp},
\end{equation}
where $\bar{S}$ is the mean 3D face shape, $\{S_{k}\vert k=1,2,\cdots, K\}$ and $\{S_{l}^{Exp}\vert l=1,2,\cdots, L\}$ are the shape and expression bases, respectively. In this paper, we employ the shape bases ($K=199$) provided by BFM \cite{Vetter_TPAMI_2003} and a simplified version of the expression bases ($L=29$) provided by FaceWarehouse \cite{Cao_TVCG_2014}.

A facial depth image is a projection of a 3D face onto 2D plane, in which the pixel values are the depth values of the corresponding 3D face vertices. In this paper, we use a weak perspective projection to approximate this 3D-to-2D projection, which is defined by scaling, rotation and translation parameters. We call these parameters (totally seven parameters) together as pose parameters.

Given a facial depth image, the goal of the 3D face reconstruction network is to estimate the aforementioned $199$-dim shape parameters, $29$-dim expression parameters and $7$-dim pose parameters, denoted as $\alpha$, $\beta$ and $\theta$, respectively. To this end, we implement a deep convolutional neural network (DCNN) to regress the parameters from the HHA image constructed from the input depth image. Specifically, we employ the SphereFace network \cite{Liu_CVPR_2017} as the base network whose last fully-connected (FC) layer is adapted to output a $235$-dim vector corresponding to the parameters to be estimated.

\subsubsection{Normalized Depth Image Generation} With the reconstructed 3D face, we next generate a facial depth image of frontal pose and neutral expression. This is fulfilled by first generating a frontal 3D face with neutral expression via substituting the reconstructed shape parameters to the 3DMM in Eq. (\ref{eq:3dmm}) (while setting the expression parameters to zero), and then applying a pre-specified 3D-to-2D projection to the 3D face. In this paper, we fit weak perspective projections to the frontal depth images and their corresponding frontal 3D faces in the training dataset based on their annotated facial landmarks, and the resulting mean weak perspective projection is used here. We call the obtained depth images as $deeply\ normalized\ depth\ images$.

\subsection{Feature Extraction Network}
Feature extraction network takes the deeply normalized depth image as input and extracts from it a robust feature representation, for face recognition. In this paper, we implement a feature extraction network based on the LightCNN model \cite{Wu_CS_2015}. LightCNN is a well-known model in the field of 2D face recognition for its simple but effective network structure. 

\subsection{Implementation Detail}

%
%

\subsubsection{Loss Functions}
Two loss functions are used in training the proposed depth-image-based face recognition method. One is an $L2$ loss defined over the normalization network, measuring the mean squared error between the estimated values and the ground truth values of the 3DMM parameters. Note that the pose, shape and expression parameters are separately normalized such that their values are of similar order of magnitude.

The other is an identification loss (i.e., softmax loss) defined over the feature extraction network. Minimization of this identification loss aims to enable the feature extraction network to generate feature representations that are effective in distinguishing different persons based on their facial depth images.

%

\subsubsection{Training Data}
The normalization network and the feature extraction network are sequentially trained with $L2$ loss and softmax loss, respectively. The training data are from three sources, the 300W-LP dataset, the Lock3DFace database, and the FRGC database. The 300W-LP dataset is used to pre-train the normalization network, the Lock3DFace is used to fine-tune the pre-trained normalization network, and the FRGC network is used to fine-tune the original LightCNN model. Faces in the depth images from these databases are detected, and the cropped face regions are resized to $128\times 128$ pixels.

The 300W-LP dataset \cite{Zhu_CVPR_2016} contains RGB images with varying poses and expressions of $3,837$ subjects as well as their corresponding ground truth 3DMM parameters. We generate depth images from the 3D faces of these subjects defined by the 3DMM parameters. The data of $3,717$ subjects are randomly chosen to pre-train the normalization network. To further augment the training data, we down-sample the depth images, add noise to them, and simulate occlusion on them. Finally, we obtain around $40$ depth images per subject.

The Lock3DFace database \cite{Zhang_IEEE_2016} consists of RGB and depth images of $509$ subjects with variations in pose, expression, illumination and occlusion. After excluding some very low quality data, we have $446$ subjects. We choose $280$ of these subjects, and calculate the ground truth shape parameter values for each subject based on his/her RGB images by using the method in \cite{Tian_FGW_2018}, which can estimate the shape parameters from a set of RGB images. As a result, all the depth images of one subject share the same shape parameters. As for the computation of the ground truth expression and pose parameter values of a depth image, we employ the method in \cite{Zhu_CVPR_2015} to fit 3DMM to the RGB image corresponding to the depth image. 


The FRGC database \cite{FRGC} contains high-precision 3D faces of $466$ subjects. We generate depth images from these 3D faces, and use the depth images to fine-tune the original LightCNN model. It is worth mentioning that in the following experiments we fine-tune the LightCNN model only using the FRGC database, though the recognition experiments are done on the Lock3DFace database.


\section{Experimental Results}
\subsection{Evaluation Protocols}
We evaluate the proposed method from two aspects: the accuracy of 3D face reconstruction and the accuracy of face recognition. For 3D face reconstruction accuracy evaluation, the data of $120$ subjects in 300W-LP are used. These subjects are different from the subjects in training dataset. We assess the 3D face reconstruction accuracy by measuring the difference between the reconstructed 3D face shape and its ground truth in terms of root mean squared error (RMSE). Given the estimated and ground truth 3DMM parameters of a test sample, the parameters are first substituted into Eq. (\ref{eq:3dmm}) to generate the corresponding 3D face shapes, and the RMSE on the test set is then calculated by
\begin{equation}
\texttt{RMSE} = \frac{1}{N_{T}}\sum_{j=1}^{N_{T}}(\| S^{*}_{j}-\hat{S}_{j} \|/n),
\end{equation} 
where $S^{*}_{j}$ and $\hat{S}_{j}$ are the ground truth and estimated 3D face shapes of the $j^{\texttt{th}}$ test sample, $n$ is the number of vertices in the 3D face shapes, and $N_{T}$ is the total number of test samples.

As for face recognition accuracy evaluation, we do face identification experiments by using the data of the remaining $166$ subjects in the Lock3DFace database as test data. In the experiments, the gallery is composed by one frontal facial depth image with neutral expression of each of the test subjects, and the probe includes two parts: depth images of frontal pose and different expressions (denoted as $Probe\_Set\_1$) and depth images of different poses (denoted as $Probe\_Set\_2$). We calculate the similarity between gallery and probe depth images based on the $cosine$ distance between their feature representations extracted by our proposed method. We report the face recognition accuracy in terms of Rank-1 identification rate.

\begin{figure}
\setlength{\belowcaptionskip}{-0.1cm}
\centering
\includegraphics[scale=0.35]{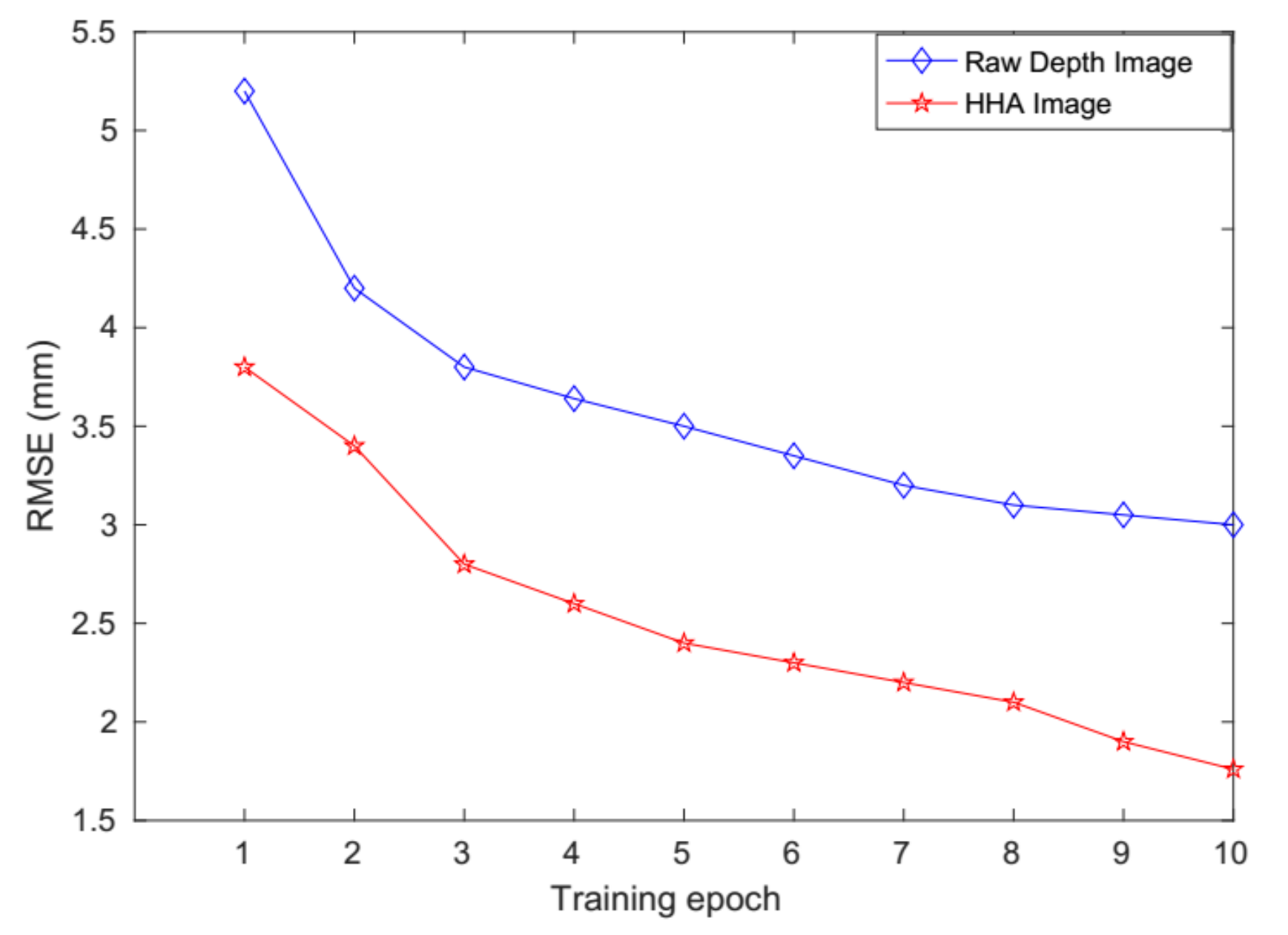}
\caption{Reconstruction accuracy in terms of RMSE w.r.t. the number of epochs. HHA images obtain obviously better reconstruction accuracy than raw depth images.}
\label{fig:3dreconstructionaccuracy}
\end{figure}

\subsection{3D Face Reconstruction Accuracy}
\begin{figure}
\setlength{\belowcaptionskip}{-0.1cm}
\centering
\includegraphics[width=0.8\linewidth]{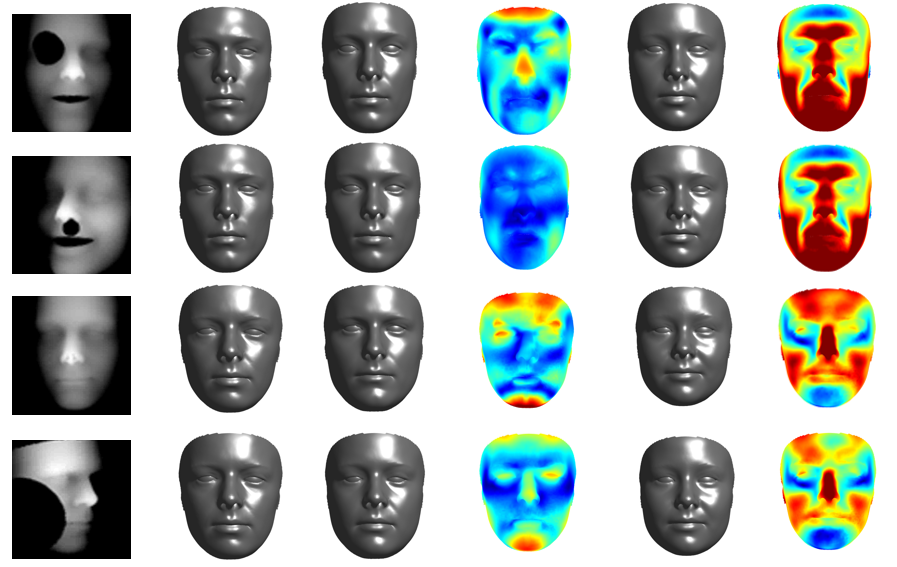}
\caption{Reconstruction results of some depth images. Each row is for one depth image. From left to right columns: Input raw probe images, ground truth 3D faces, reconstructed 3D faces using HHA images and the corresponding error maps, and reconstructed 3D faces using raw depth images and the corresponding error maps. Errors increase as the color changes from dark blue to dark red in the error maps. Note that occlusions are simulated on the test depth images.}
\label{fig:3dreconstructionexample}
\end{figure}

Figure \ref{fig:3dreconstructionaccuracy} gives the 3D face reconstruction accuracy (in terms of RMSE) with respect to the number of training epochs of our proposed method on the test data in 300W-LP. In order to show the effectiveness of HHA images, we report the results of using raw depth images as the input to the 3D face reconstruction DCNN. As can be seen, HHA images achieve obviously lower reconstruction errors than raw depth images. Figure \ref{fig:3dreconstructionexample} shows some example reconstruction results.


\begin{table}[h]
\setlength{\tabcolsep}{4pt} 
\begin{center}
\label{table:fraccuracy}
\caption{Rank-1 identification rates on the Lock3DFace database.}
\begin{tabular}{c|c|c|c}
\hline
\textbf{}	&$Probe\_Set\_1$	&$Probe\_Set\_2$	 &$Mean$	\\
\hline
Raw Depth        &$62\%$  &$10.5\%$   &$36.25\%$    \\
\hline
Ours 	&$\textbf{92\%}$	 &$\textbf{80\%}$	 &$\textbf{86\%}$	 \\
\hline
Baseline\cite{Zhang_IEEE_2016} 	&$74.12\%$	 &$18.63\%$	 &$46.38\%$	 \\
\hline
\end{tabular}
\end{center}
\vspace{-.5cm}
\end{table}

\begin{figure}[h]
\centering
\includegraphics[width=0.8\linewidth]{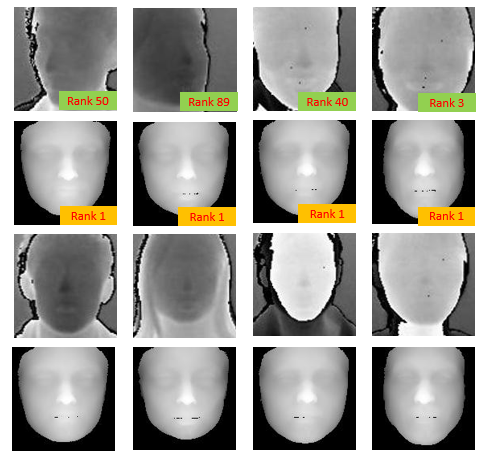}
\caption{Recognition results of some probe depth images. Each column is for a probe depth image. From top to bottom rows: Input raw probe depth images, Deeply normalized probe depth images, Corresponding gallery depth images, Deeply normalized gallery depth images. The ranks at which the corresponding gallery depth images are hit are shown on the right bottom of the probe depth images.}
\label{fig:facerecognitionexample}
\end{figure}

\subsection{Face Recognition Accuracy}

Face recognition accuracy is presented in Table $1$ in terms of rank-1 identification rate. From these results, the following observations can be made. (i) When recognizing frontal faces with varying expressions, our proposed method improves the accuracy by $30\%$ compared with using the raw depth images. This proves the effectiveness of our method in removing expression-induced deformation in depth images. (ii) As for depth images of arbitrary poses, our method makes a more substantial improvement. A possible reason is that our method can well generate the frontal view depth images while recovering the invisible regions. (iii) On average, our proposed method significantly advances the state-of-the-art performance on the Lock3DFace database, thanks to the effective normalization of the depth images in favor of face recognition. See Fig. \ref{fig:facerecognitionexample} for the recognition results of some probe depth images.

\section{Conclusions}
This paper has proposed a novel method for depth-to-depth face recognition. It employs a deep neural network to reconstruct 3D faces from single depth images under arbitrary poses and expressions, and then generates pose-and-expression normalized facial depth images. This way, factors like noisy depth values and varying poses and expressions that may distract face recognition are suppressed. The proposed method then utilizes another deep neural network to extract robust features from the deeply normalized depth images. Experimental results demonstrate the effectiveness of the proposed method in recognizing faces of arbitrary poses and expressions. In the future, we are going to evaluate the proposed method on more benchmarks, and further improve the method by better preserving the identity information during depth image normalization.

\section*{Acknowledgements}
This work is supported by the National Key Research and Development Program of China (2017YFB0802300) and the National Natural Science Foundation of China (61773270).

\end{document}